\documentclass[journal=jpclcd,manuscript=letter]{achemso}

\usepackage{algorithm}
\usepackage{algpseudocode}

\usepackage[T1]{fontenc}
\usepackage{amsmath}
\usepackage{amssymb}
\usepackage{bm}
\usepackage{graphicx}
\usepackage{float}
\usepackage{chemformula}
\usepackage{booktabs}

\author{Aaron L. Feller}
\email{aaron.feller@utexas.edu}
\affiliation[UT Austin]{Interdisciplinary Life Sciences, The University of Texas at Austin, Austin, TX}
\alsoaffiliation[Novo Nordisk]{Molecular AI, Novo, Lexington, MA}

\author{Kristine Deibler}
\affiliation[Novo Nordisk]{Molecular AI, Novo, Lexington, MA}

\author{Maxim Secor}
\affiliation[Novo Nordisk]{Molecular AI, Novo, Lexington, MA}

\title{EnsembleEGNN: Set-Based Graph Learning for Thermodynamic Ensembles of Cyclic Peptides}



\begin{tocentry}
    \centering
    \includegraphics[width=3.25in,height=1.75in,keepaspectratio]{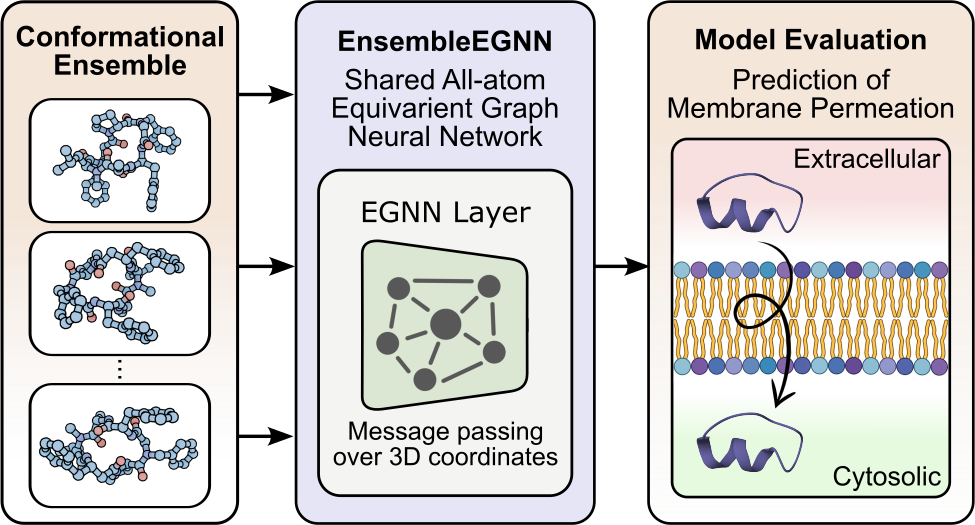}
\end{tocentry}

\begin{document}

\begin{abstract}
    Molecular graph encoding often relies on a single, static structure, ignoring the thermodynamic ensemble of molecules that are present in solution. Here, we introduce EnsembleEGNN, a foundation model that encodes structural ensembles by processing individual conformers through shared equivariant graph neural network layers, pooled with a set attention block, to make property predictions from the whole ensemble. Pretrained on the CREMP cyclic peptide dataset using multi-task self-supervision, the model is trained to encode the conformational variability of each molecule. When predicting membrane permeability from the CycPeptMPDB benchmark, EnsembleEGNN achieves an $R^2$ of $0.477$ under random cross-validation, outperforming a sequence-only BERT baseline ($R^2=0.439$). This representation advantage persists under rigorous out-of-distribution Butina splits ($R^2=0.401$ versus $0.354$). Finally, a hybrid architecture co-training EnsembleEGNN with the BERT model achieves the highest overall accuracy across both random ($R^2=0.538$) and structural holdout evaluations ($R^2=0.444$). These results demonstrate that encoding conformational ensembles into latent representations improves predictions for properties governed by thermodynamics.
\end{abstract}

Graph foundation models have shown progress in mapping molecular structures to chemical properties~\citep{hsu2022learning, mendez2024mole, liu2025graph}. However, state-of-the-art 3D architectures typically process only a single static conformation per molecule~\cite{schutt2018schnet, gasteiger2020directional, schutt2021equivariant, zhou2023uni, otero2025pepmnet, liu2026molx, stark20223d, liao2022equiformer}, while alternative approaches omit 3D data entirely in favor of 2D connectivity graphs~\cite{gilmer2017neural, xu2018powerful, wang2022molecular, graff2026chemprop}. Both simplifications fundamentally misrepresent physical reality. In solution, molecules dynamically populate a distribution of states known as a thermodynamic ensemble~\citep{chandler1987introduction, tuckerman2023statistical}.

Using a single conformation is particularly problematic when modeling cyclic peptides, as they are inherently flexible~\citep{hui2025molecular, bayaraa2026fast, damjanovic2021elucidating}. Molecular behavior impacting key properties like membrane permeability and protease resistance may be dictated by a specific conformer within the ensemble rather than the single most stable state~\citep{yudin2015macrocycles, calvo2025computational}, often described as structural chameleonicity~\cite{whitty2016quantifying, ono2021cyclosporin}. While recent methods attempt to aggregate multiple states via naive mean pooling~\citep{zhu2024learning} or mathematically complex fragment-aware graph transformers~\citep{nguyen2026facet}, these approaches often either disregard the underlying thermodynamic context or are designed for small molecules and translate poorly to continuous macrocyclic rings. With conformer-generation pipelines improving~\citep{aranganathan2025modeling, jing2022torsional, ganea2021geomol}, the core representational challenge for these flexible scaffolds remains: given $C$ conformers of $N$ atoms with physical Boltzmann weights $(w_c)$, how can a model efficiently aggregate their geometric information into a single embedding?

\begin{figure}[ht]
    \centering
    \includegraphics[width=0.99\linewidth]{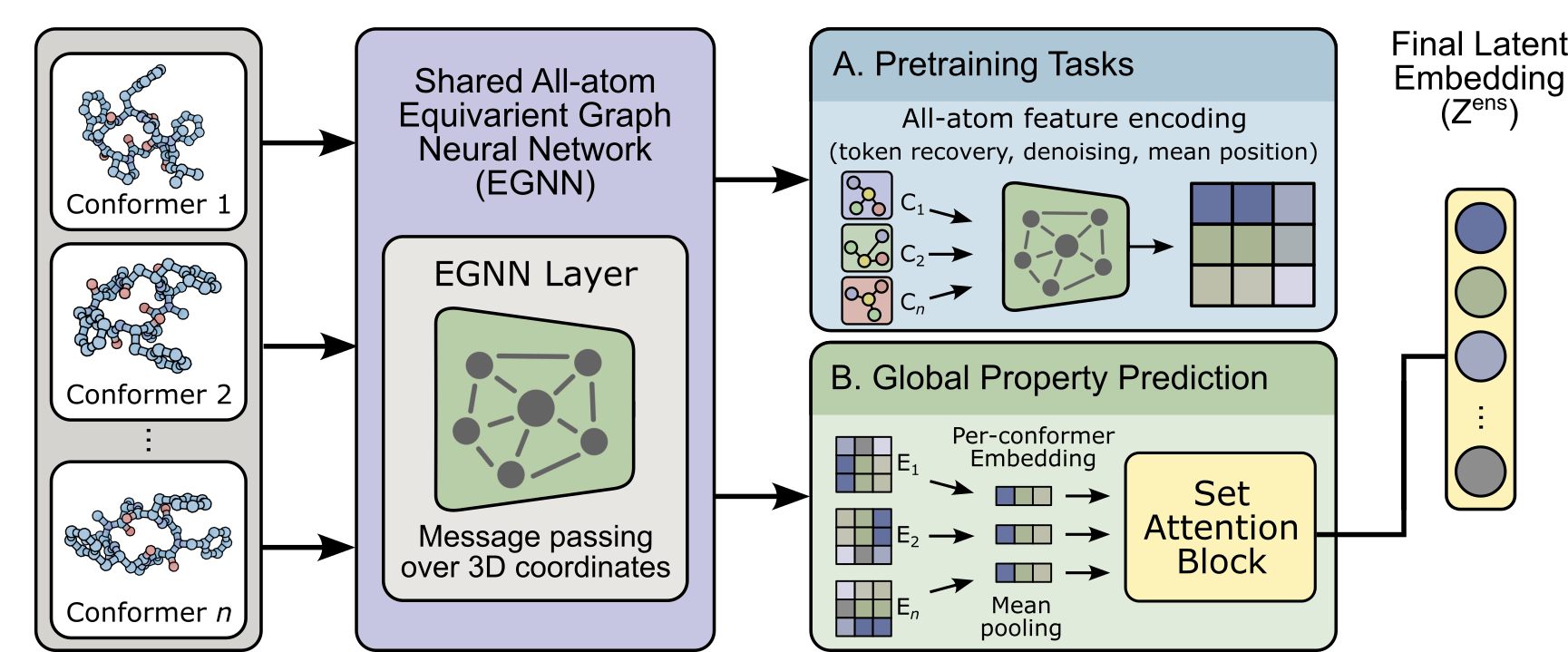}
    \caption{\textbf{Overview of the EnsembleEGNN architecture and workflows.} A conformational ensemble consisting of $C$ conformers is processed by a shared all-atom equivariant graph neural network (EGNN), which performs equivariant message passing over 3D spatial coordinates to update atomic representations. (A) Self-supervised pretraining tasks optimize all-atom feature encodings across the ensemble using masked token recovery, coordinate denoising, and mean position distance reconstruction. (B) For downstream property prediction, atom features within each conformer are aggregated via mean pooling into per-conformer embeddings ($E_1, \dots, E_n$), which are then integrated through a set attention block to produce the final latent ensemble embedding ($Z^{\mathrm{ens}}$).}
    \label{fig:diagram}
\end{figure}

\begin{figure*}[t]
  \centering
  \includegraphics[width=0.99\textwidth]{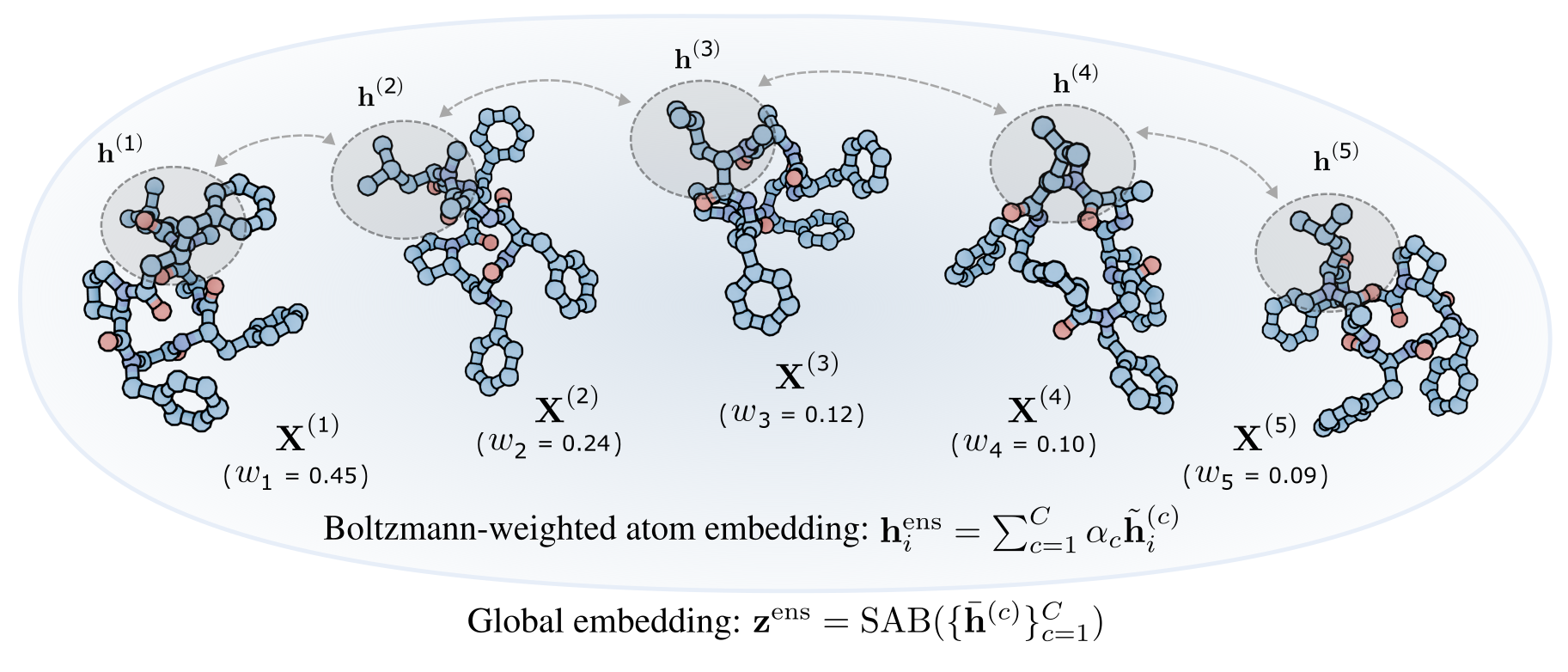}
  \caption{\textbf{Geometric variance and thermodynamic weighting across a molecular ensemble.} A cyclic peptide is represented as a distribution of $C$ distinct spatial states, each defined by its 3D coordinates ($\mathbf{X}^{(c)}$) and corresponding physical Boltzmann weight ($w_c$). Shaded regions track the local spatial environment of a single shared atom ($\mathbf{h}^{(c)}$) as it fluctuates across the accessible conformations. Rather than relying on a static structure, these conformation-specific atomic features are aggregated via a Boltzmann-informed attention mechanism ($\alpha_c$) to form a unified, flexibility-aware atom embedding ($\mathbf{h}_i^{\text{ens}}$) prior to global pooling.}
  \label{fig:ensemble}
\end{figure*}

To address this, we present EnsembleEGNN, an encoder model for cyclic peptides that processes each conformer separately with a shared equivariant graph neural network~\cite{satorras2021en} (EGNN). Treating the conformational ensemble as an unordered set~\citep{zaheer2017deep}, the model then combines those conformer-level representations into a single ensemble embedding using a set attention block~\citep{lee2019set} (SAB) (Fig.~\ref{fig:diagram}). In this view, the same ordered set of atoms is observed under multiple 3D conformations, allowing the model to explicitly capture the inherent flexibility of the molecule (Fig.~\ref{fig:ensemble}). This process allows for the model to learn prioritization of conformers that are important on a per-task basis, maintaining low architectural complexity while still learning from the full thermodynamic distribution of each molecule.

To bridge 1D token identity and 3D geometry, we represent each peptide as a set of tokens, where each conformation reuses the same atom nodes with different position coordinates. To construct these inputs from CREMP~\citep{grambow2024cremp}, we extract all available conformers, rank them by their Boltzmann weights, retain the top $C=5$ states, and normalize them to sum to one before serialization. Node (atom) tokens ($a$) correspond directly to atom identities, giving a coordinate tensor of shape $C \times a \times 3$ together with per-conformer weights.

Each conformer is processed through $L$ EGNN layers~\citep{satorras2021en}. The model updates both the node features and coordinates by passing messages exclusively over the local $k$-nearest-neighbor graph within that specific conformer:

\begin{equation}
  \mathbf{h}_i^{(c,\ell+1)}, \mathbf{x}_i^{(c,\ell+1)}
  = \mathrm{EGNNLayer}\!\left(
      \mathbf{h}_i^{(c,\ell)},\,
      \mathbf{x}_i^{(c,\ell)},\,
      \mathcal{N}_k(i, c)
    \right)
\end{equation}

where the initial node features are $\mathbf{h}_i^{(c,0)} = \mathrm{Embed}(a_i)$, the initial coordinates are $\mathbf{x}_i^{(c,0)} = \mathbf{X}_i^{(c)}$, and $\mathcal{N}_k(i,c)$ denotes the $k$ nearest spatial neighbors of node $i$ in conformer $c$. The choice of EGNN layers ensures that the per-conformer message passing is strictly equivariant to rotations and translations in 3D space. EnsembleEGNN uses all-atom representations with 12 nearest neighbors.

Let $\tilde{\mathbf{h}}^{(c)} \in \mathbb{R}^{N \times d}$ denote the final post-EGNN node (i.e., atom) features for conformer $c$. To pool these separate structures into a unified representation, each conformer receives an attention weight driven by two signals: a learned score from the mean ensemble embedding and a thermodynamic prior from the Boltzmann weight. We compute these attention weights as:

\begin{equation}
  \alpha_c = \frac{\exp\!\left(f_\phi(\bar{\mathbf{h}}^{(c)}) + \log w_c\right)}
  {\sum_{c'}\exp\!\left(f_\phi(\bar{\mathbf{h}}^{(c')}) + \log w_{c'}\right)}
\end{equation}

where $\bar{\mathbf{h}}^{(c)} = \frac{1}{N}\sum_i \tilde{\mathbf{h}}_i^{(c)}$ is the mean node feature for conformer $c$ and $f_\phi$ is a 2-layer MLP. Using these attention weights, we compute two distinct ensemble embeddings:

\begin{equation}
  \mathbf{h}_i^{\mathrm{ens}} = \sum_{c=1}^C \alpha_c\,\tilde{\mathbf{h}}_i^{(c)},
  \qquad
  \mathbf{z}^{\mathrm{ens}} = \mathrm{SAB}\!\left(\{\bar{\mathbf{h}}^{(c)}\}_c\right)
\end{equation}

The first term, $\mathbf{h}_i^{\mathrm{ens}}$, is a per-atom embedding obtained by taking the attention-weighted average of each atom across all conformers. The second term, $\mathbf{z}^{\mathrm{ens}} \in \mathbb{R}^d$, is a global graph-level embedding generated by passing the conformer embeddings through a set attention block (SAB)~\citep{lee2019set}. Utilizing a set attention block with $M$ inducing points reduces the conformer fusion complexity from $\mathcal{O}(C^2)$ with standard pairwise attention to $\mathcal{O}(CM)$, enabling efficient scaling across large conformational ensembles. The complete pseudocode for the conformer-aware forward pass is provided in Algorithm S1 of the Supporting Information.

During pretraining, we corrupt both the node identities and the coordinates. We randomly select 15\% of valid node positions for masked-token supervision. Following a BERT-style corruption scheme, 80\% of those selected positions are replaced with a mask token, 10\% are replaced with a random token, and 10\% are left unchanged. We also add Gaussian noise to valid node and conformer positions, producing a noisy coordinate tensor that the model must refine back toward the clean structure.

The composite pretraining loss is:
\begin{equation}
  \mathcal{L} = 0.3 (\mathcal{L}_{\mathrm{tok}})
              + 0.5 (\mathcal{L}_{\mathrm{coord}})
              + 0.2 (\mathcal{L}_{\mathrm{dist}})
\end{equation}

The masked-token term $(\mathcal{L}_{\mathrm{tok}})$ is a cross-entropy loss computed at the masked valid nodes. The coordinate term $(\mathcal{L}_{\mathrm{coord}})$ is an MSE loss between the final EGNN-refined coordinates and the clean conformer coordinates. The distance term $(\mathcal{L}_{\mathrm{dist}})$ compares pairwise distances computed from projected per-atom embeddings with pairwise distances computed from a Boltzmann-weighted mean coordinate set. 

\begin{figure}[ht]
  \centering
  \includegraphics[width=0.99\columnwidth]{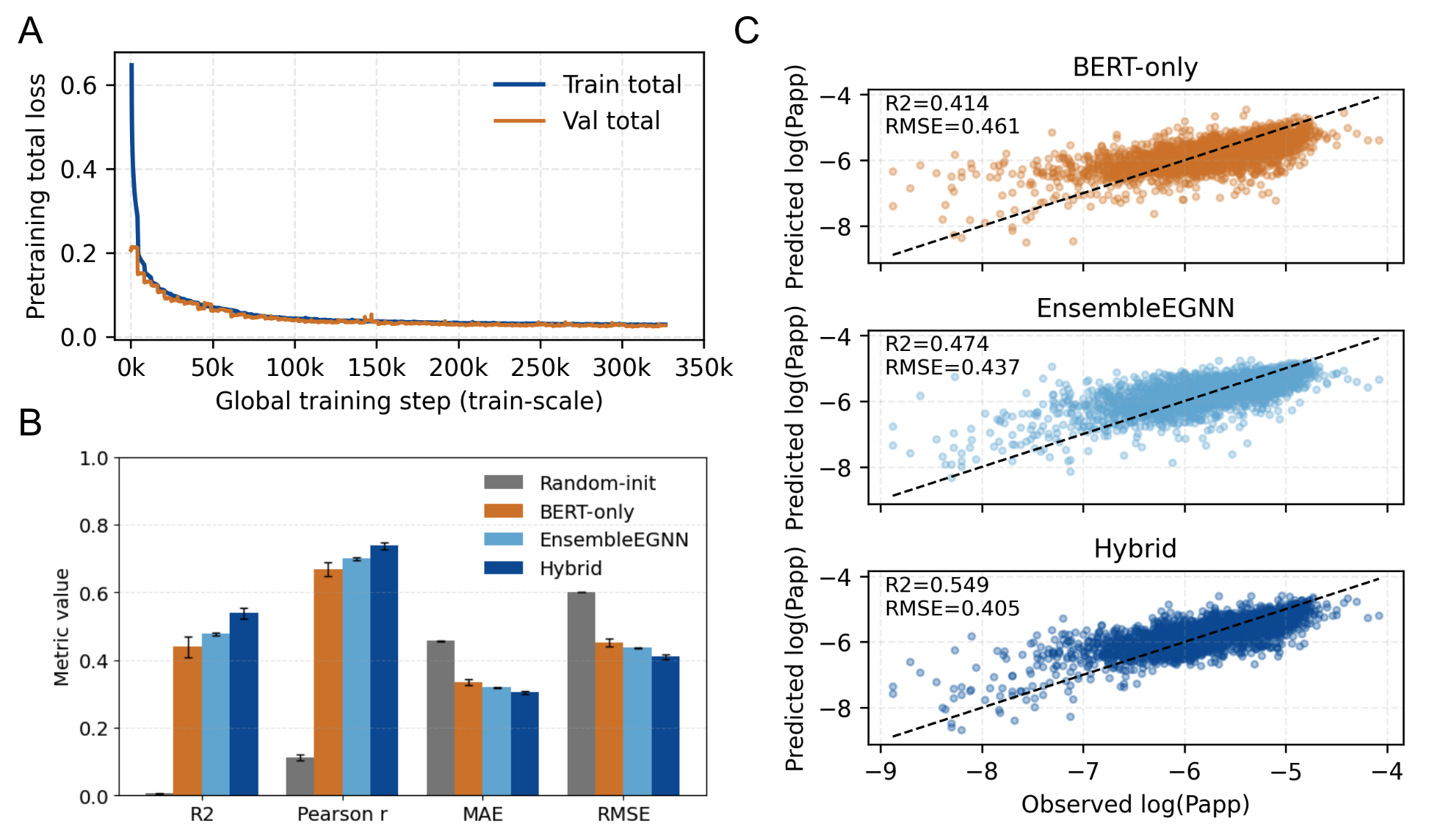}
  \caption{\textbf{Pretraining convergence and cross-validation predictive performance.} (A) Trajectories of the training loss (50-step moving average) and validation loss plotted against global training steps, illustrating stable model convergence on the CREMP dataset. (B) Comparison of aggregate mean evaluation metrics for the randomly initialized geometric baseline, sequence-only BERT, pretrained EnsembleEGNN, and co-trained Hybrid architectures on the held-out PAMPA dataset. Error bars denote the standard error across three independent training runs. (C) Parity plots comparing predicted versus experimental $\log(P_{\text{app}})$ values across the complete evaluation set ($n=2979$). Proximity to the diagonal dashed identity line illustrates prediction accuracy, highlighting the tighter correlation and lower error achieved by the Hybrid model.}
  \label{fig:metrics}
\end{figure}

During self-supervised pretraining, we log train and validation total loss throughout optimization and track convergence on a train-scaled global-step axis. These learning dynamics showed a smoothed train-loss trajectory and validation loss decreasing with training loss (Fig.~\ref{fig:metrics}A).

For quantitative downstream evaluation, we used CREMP-CycPeptMPDB~\cite{grambow2024cremp}, which contains cyclic peptide conformers paired with artificial membrane permeation scores, yielding $n=3003$ labeled measurements before boundary filtering. We removed exact boundary targets at $\log(P_{\text{app}})$=-10.0, -8.0, and -4.0, resulting in a final evaluation set of $n=2979$. We used fixed random 5-fold cross-validation and report fold-wise and aggregate metrics over all held-out examples. We additionally constructed an out-of-distribution (OOD) training split utilizing Butina clustering~\citep{butina1999unsupervised} on Morgan fingerprints~\cite{morgan1965generation}. This method assigns structural groups to individual folds, reducing data leakage and creating a better evaluation for how well the model generalizes (detailed in Table S1 of the Supporting Information).

We compared four model-types under identical random splits: BERT-only, a SMILES baseline utilizing the PeptideCLM-2 encoder \cite{feller2026scaling}; EnsembleEGNN-random-init, the geometric model trained from random initialization without pretraining; EnsembleEGNN, the pretrained geometric model; and Hybrid, the full co-training EnsembleEGNN and PeptideCLM-2 with concatenated embeddings.

\begin{table*}[ht]
  \centering
  \caption{\textbf{Model performance metrics under random-fold cross-validation.} Predictive accuracy on held-out CREMP-CycPeptMPDB data ($n=2979$). Values are reported as the mean $\pm$ standard error across 3 independent training replicates.}
  \label{tab:performance_metrics}
  \small
  \begin{tabular}{lcccc}
    \toprule
    \textbf{Model} & \bm{$R^2$} & \textbf{Pearson} \bm{$r$} & \textbf{MAE} & \textbf{RMSE} \\
    \midrule
    Hybrid & \textbf{0.538 $\pm$ 0.029} & \textbf{0.737 $\pm$ 0.016} & \textbf{0.304 $\pm$ 0.008} & \textbf{0.409 $\pm$ 0.013} \\
    EnsembleEGNN & 0.477 $\pm$ 0.009 & 0.699 $\pm$ 0.009 & 0.319 $\pm$ 0.002 & 0.436 $\pm$ 0.004 \\
    BERT-only & 0.439 $\pm$ 0.053 & 0.667 $\pm$ 0.035 & 0.334 $\pm$ 0.014 & 0.451 $\pm$ 0.022 \\
    Random-init & 0.005 $\pm$ 0.004 & 0.113 $\pm$ 0.016 & 0.457 $\pm$ 0.003 & 0.601 $\pm$ 0.001 \\
    \bottomrule
  \end{tabular}
\end{table*}

\begin{table}[ht]
  \centering
    \caption{\textbf{Detailed fold-wise \bm{$R^2$} evaluation.} Comparison of predictive performance across five distinct test folds. Results represent the mean $\pm$ standard error of three independent replicates.}
  \label{tab:folds}
  \small
  \begin{tabular}{lccc}
    \toprule
    \textbf{Fold} & \textbf{BERT-only} & \textbf{EnsembleEGNN} & \textbf{Hybrid} \\
    \midrule
    1 & 0.426 $\pm$ 0.019 & 0.503 $\pm$ 0.007 & \textbf{0.565 $\pm$ 0.011} \\
    2 & 0.472 $\pm$ 0.026 & 0.456 $\pm$ 0.012 & \textbf{0.494 $\pm$ 0.035} \\
    3 & 0.430 $\pm$ 0.061 & 0.509 $\pm$ 0.031 & \textbf{0.553 $\pm$ 0.018} \\
    4 & 0.462 $\pm$ 0.057 & 0.451 $\pm$ 0.016 & \textbf{0.526 $\pm$ 0.054} \\
    5 & 0.402 $\pm$ 0.060 & 0.460 $\pm$ 0.012 & \textbf{0.552 $\pm$ 0.018} \\
    \bottomrule
  \end{tabular}
\end{table}

Under random-fold cross-validation, overall performance on the PAMPA dataset details the necessity of the pretrained conformer-aware approach (Table~\ref{tab:performance_metrics}). While the randomly initialized EnsembleEGNN performed near random ($R^2 = 0.005$), pretraining the model on CREMP outperformed the sequence-only BERT baseline ($R^2 = 0.477$ vs. $0.439$). The co-trained Hybrid model ultimately achieved the strongest aggregate metrics, improving the final $R^2$ to $0.538$, resulting in a Pearson $r$ of $0.737$. The Hybrid model consistently achieved the strongest mean $R^2$ performance across all 5 individual cross-validation folds (Table~\ref{tab:folds}). Visualizations of the overall metrics further demonstrate this performance gap (Fig.~\ref{fig:metrics}B), supported by per-model held-out prediction scatter plots (Fig.~\ref{fig:metrics}C). 

In addition to evaluating the prior methods, we also conducted an ablation study on the structural OOD splits to isolate the impact of the conformer pooling mechanism. Specifically, we compared the set attention fusion against direct Boltzmann pooling, a learned pooling, and a baseline using only the single highest-weighted conformer.

\begin{table*}[t]
  \centering
  \caption{\textbf{Conformer-pooling ablation under structural OOD evaluation.} Five-fold cross-validation on the $n=2979$ boundary-filtered measurements with whole Butina analog groups assigned to individual folds. Values are mean $\pm$ standard error across three training replicates.}
  \label{tab:ood_ablation}
  \small
  \begin{tabular}{lcccc}
    \toprule
    \textbf{Model} & \bm{$R^2$} & \textbf{Pearson} \bm{$r$} & \textbf{MAE} & \textbf{RMSE} \\
    \midrule
    Hybrid SAB+BERT & \textbf{0.444 $\pm$ 0.010} & \textbf{0.675 $\pm$ 0.005} & \textbf{0.335 $\pm$ 0.003} & \textbf{0.449 $\pm$ 0.004} \\
    EnsembleEGNN & 0.401 $\pm$ 0.005 & 0.647 $\pm$ 0.002 & 0.339 $\pm$ 0.002 & 0.466 $\pm$ 0.002 \\
    BERT-only & 0.354 $\pm$ 0.048 & 0.601 $\pm$ 0.034 & 0.360 $\pm$ 0.015 & 0.483 $\pm$ 0.018 \\
    Top-1 conformer & 0.349 $\pm$ 0.017 & 0.619 $\pm$ 0.005 & 0.347 $\pm$ 0.004 & 0.486 $\pm$ 0.006 \\
    Boltzmann pooling & 0.336 $\pm$ 0.019 & 0.618 $\pm$ 0.004 & 0.350 $\pm$ 0.001 & 0.491 $\pm$ 0.007 \\
    Learned pooling & 0.335 $\pm$ 0.020 & 0.611 $\pm$ 0.011 & 0.353 $\pm$ 0.004 & 0.491 $\pm$ 0.007 \\
    \bottomrule
  \end{tabular}
\end{table*}

Among the geometric models, full set attention fusion achieved the highest mean structural-OOD performance ($R^2=0.401 \pm 0.005$) (Table~\ref{tab:ood_ablation}). It exceeded both direct Boltzmann pooling ($R^2=0.336 \pm 0.019$) and learned Boltzmann-weighted pooling without set attention ($R^2=0.335 \pm 0.020$), indicating that the gain is not explained by a weighted average of conformers alone. Restricting the model to the single highest-Boltzmann-weight conformer also reduced performance ($R^2=0.349 \pm 0.017$), consistent with task-relevant information being distributed across the ensemble. Finally, combining the set attention geometric representation with the sequence encoder yielded the best OOD result ($R^2=0.444 \pm 0.010$), improving on both the geometric-only model and BERT-only baseline. These results support complementary use of both sequence-based and ensemble-based molecular representations.

The conformer aggregation problem is not unique to peptides. Protein loop flexibility~\citep{marks2018predicting}, RNA secondary-structure ensembles~\citep{bauskar2025boltzmann}, and small-molecule tautomers~\citep{greenwood2010towards} all involve dynamic structural states. Our results suggest that explicitly learning from multiple conformers, rather than relying on a single structure, can be a useful design principle for molecular foundation models.

Our approach has three primary limitations. Architecturally, the current EGNN layers perform message passing only within each conformer; interactions across conformers are introduced later through conformer pooling rather than through explicit cross-conformer message passing. Data-wise, this study was strictly bounded by the coverage and accuracy of the precomputed CREMP ensembles. Finally, while our hybrid model outperforms the sequence baseline on this benchmark, the CREMP-CycPeptMPDB dataset remains a low-data regime ($n=2979$), which warrants caution when making generalized cross-paper comparisons. Future work will explore richer cross-conformer interaction mechanisms and integrate scalable, high-fidelity conformer generation to expand beyond existing library constraints.

In summary, EnsembleEGNN demonstrates that explicitly encoding a molecular ensemble with shared EGNN layers and pooling conformer representations into a single thermodynamically informed embedding outperforms static-structure baselines. Through multi-task self-supervised pretraining on CREMP, the model learns to recover masked tokens, refine noisy coordinates, and preserve coarse ensemble geometry. With transfer learning to the CREMP-CycPeptMPDB benchmark, EnsembleEGNN outperforms a pretrained BERT architecture across both random cross-validation and rigorous structural out-of-distribution splits. Furthermore, co-training these representations in a hybrid model yields the highest overall predictive accuracy in all evaluation settings. More broadly, this work shows that directly learning from conformer ensembles can significantly improve cyclic-peptide representation learning.

In order to ensure reproducibility and allow the scientific community to build on this concept, we release the model architecture, pretrained checkpoint, and training code, as well as generation of manuscript figures for this paper at https://github.com/AaronFeller/EnsembleEGNN.

\begin{suppinfo}
The Supporting Information is available free of charge at \textit{[URL to be inserted by ACS]}. 

Pseudocode for the conformer-aware forward pass (Algorithm S1) and comprehensive model architecture and pretraining/finetuning hyperparameters (Table S1).
\end{suppinfo}

\begin{acknowledgement}
A.L.F. would like to thank the team at Novo for supporting open research, which has led to the completion of this manuscript and the release of all code under MIT license. LLMs were used during the preparation of the manuscript solely as writing aids to catch typographical and grammatical errors and to improve writing style and latex formatting.
\end{acknowledgement}

\bibliography{references}

\clearpage
\setcounter{table}{0}
\renewcommand{\thetable}{S\arabic{table}}

\setcounter{algorithm}{0}
\renewcommand{\thealgorithm}{S\arabic{algorithm}}

\begin{algorithm}[h]
   \caption{EnsembleEGNN Conformer-Aware Fusion Forward Pass}
   \label{alg:fusion}
\begin{algorithmic}[1]
   \State {\bfseries Input:} Peptide with $N$ atoms having: 
   \Statex \hspace{1.1cm} atom types $\{a_i\}_{i=1}^N$,
   \Statex \hspace{1.1cm} per-conformer atom coordinates $\{\mathbf{X}^{(c)}\}_{c=1}^C$, 
   \Statex \hspace{1.1cm} and Boltzmann weights $\{w_c\}_{c=1}^C$
   \State Initialize atom features $\mathbf{h}_i^{(c,0)} = \mathrm{Embed}(a_i)$
   \For{$\ell = 0$ {\bfseries to} $L-1$}
       \State Update $\mathbf{h}_i^{(c,\ell+1)}, \mathbf{x}_i^{(c,\ell+1)}$ via EGNN 
   \EndFor
   \State Let $\tilde{\mathbf{h}}^{(c)} = \mathbf{h}^{(c,L)}$ denote final atom features
   \For{$c = 1$ {\bfseries to} $C$}
       \State Conformer token: $\bar{\mathbf{h}}^{(c)} = \frac{1}{N} \sum_{i=1}^N \tilde{\mathbf{h}}_i^{(c)}$
       \State Attention logit: $u_c = f_\phi(\bar{\mathbf{h}}^{(c)}) + \log w_c$
   \EndFor
   \State Attention weights: $\alpha_c = \frac{\exp(u_c)}{\sum_{c'} \exp(u_{c'})}$
   \State Ensemble atoms: $\mathbf{h}_i^{\mathrm{ens}} = \sum_{c=1}^C \alpha_c \tilde{\mathbf{h}}_i^{(c)}$
   \State Global embedding: $\mathbf{z}^{\mathrm{ens}} = \mathrm{SAB}(\{\bar{\mathbf{h}}^{(c)}\}_{c=1}^C)$
   \State {\bfseries Output:} $\mathbf{h}^{\mathrm{ens}}$ and $\mathbf{z}^{\mathrm{ens}}$
\end{algorithmic}
\end{algorithm}

\newpage

\begin{table}[h]
  \centering
  \caption{\textbf{EnsembleEGNN architecture and training hyperparameters.}}
  \label{tab:hyperparameters}
  \small
  \resizebox{\columnwidth}{!}{
  \begin{tabular}{llc}
    \toprule
    \textbf{Category} & \textbf{Parameter} & \textbf{Value} \\
    \midrule
    \textbf{EGNN backbone} & Parameters & 7.85M \\
    & Number of EGNN layers ($L$) & 6 \\
    & Effective hidden dimension ($d$) & 448 \\
    & Number of neighbors ($k$) & 12 \\
    & Dropout & 0.1 \\
    \midrule
    \textbf{Conformer fusion} & Parameters & 3.42M \\
    & Fusion mode & \texttt{setattn} \\
    & Set-attention blocks & 2 \\
    & Inducing points & 4 \\
    & Attention heads & 8 \\
    & Conformer-prior bias & enabled \\
    \midrule
    \textbf{Pretraining: EnsembleEGNN} & Epochs & 80 \\
    & Batch size & 8 \\
    & Learning rate & $2 \times 10^{-4}$ \\
    & Weight decay & $1 \times 10^{-4}$ \\
    & Masking ratio & 15\% \\
    & Corruption probabilities (mask/random/keep) & 0.8 / 0.1 / 0.1 \\
    & Coordinate noise std ($\sigma$) & 0.15~\AA \\
    & Loss weights ($\lambda_{\mathrm{tok}}$, $\lambda_{\mathrm{coord}}$, $\lambda_{\mathrm{dist}}$) & 0.3 / 0.5 / 0.2 \\
    \midrule
    \textbf{Fine-tuning: EnsembleEGNN} & Epochs & 20 \\
    & Batch size & 8 \\
    & Learning rate & $1 \times 10^{-4}$ \\
    & Weight decay & $1 \times 10^{-4}$ \\
    & Regression head layers & 2 \\
    & Head dropout & 0.1 \\
    & Early stopping patience & 3 \\
    \midrule
    \textbf{Fine-tuning: BERT} & Encoder & \texttt{aaronfeller/peptideclm-2-hybrid-base} \\
    & Epochs & 20 \\
    & Batch size & 16 \\
    & Learning rate & $3 \times 10^{-4}$ \\
    & Weight decay & $1 \times 10^{-2}$ \\
    & Regression head layers & 2 \\
    & Head dropout & 0.1 \\
    & Early stopping patience & 3 \\
    \midrule
    \textbf{Fine-tuning: Hybrid} & Geometric init checkpoint & \texttt{best\_pretrain.pt} \\
    & Epochs & 20 \\
    & Batch size & 8 \\
    & Learning rates (head / ensemble / CLM) & $2 \times 10^{-4}$ / $1 \times 10^{-4}$ / $2 \times 10^{-5}$ \\
    & Weight decay & $1 \times 10^{-4}$ \\
    & Regression head layers & 2 \\
    & Head dropout & 0.1 \\
    & Early stopping patience & 3 \\
    \bottomrule
  \end{tabular}
  }
\end{table}

\end{document}